\documentclass[pdflatex,sn-mathphys]{sn-jnl}

\jyear{2021}%

\theoremstyle{thmstyleone}%
\theoremstyle{thmstyletwo}%

\theoremstyle{thmstylethree}%

\usepackage{aas_macros}
\def\bi#1{\hbox{\boldmath{$#1$}}}

\begin{document}

\title[DLO with NFs]{Deterministic Langevin Unconstrained Optimization with Normalizing Flows}


\author*[1,2]{\fnm{James M.} \sur{Sullivan}}\email{jmsullivan@berkeley.edu}

\author[2,3,4]{\fnm{Uro\v{s}} \sur{Seljak}}\email{useljak@berkeley.edu}

\affil*[1]{\orgdiv{Berkeley Center for Cosmological Physics}, \orgname{Campbell Hall}, \orgaddress{\street{University Dr.}, \city{Berkeley}, \postcode{94720}, \state{CA}, \country{U.S.A.}}}

\affil[2]{\orgdiv{Department of Astronomy},  \orgname{Campbell Hall}, \orgaddress{\street{University Dr.}, \city{Berkeley}, \postcode{94720}, \state{CA}, \country{U.S.A.}}}

\affil[3]{\orgdiv{Physics Division},  \orgname{Lawrence Berkeley National Laboratory}, \orgaddress{\street{1 Cyclotron Road}, \city{Berkeley}, \postcode{94720}, \state{CA}, \country{U.S.A.}}}

\affil[4]{\orgdiv{Department of Physics},  \orgname{Campbell Hall}, \orgaddress{\street{University Dr.}, \city{Berkeley}, \postcode{94720}, \state{CA}, \country{U.S.A.}}}

\abstract{We introduce a global, gradient-free surrogate optimization strategy for expensive black-box functions inspired by the Fokker-Planck and Langevin equations. 
These can be written as an optimization problem where the objective is the target function to maximize minus the logarithm of the current density of evaluated samples. This 
objective balances exploitation of the target objective with exploration of low-density regions.
The method, Deterministic Langevin Optimization (DLO), relies on a Normalizing Flow density estimate to perform active learning and select proposal points for evaluation.
This strategy differs qualitatively from the widely-used acquisition functions employed by Bayesian Optimization methods, and can accommodate a range of surrogate choices.
We demonstrate superior or competitive progress toward objective optima on standard synthetic test functions, as well as on non-convex and multi-modal posteriors of moderate dimension. 
On real-world objectives, such as scientific and neural network hyperparameter optimization, DLO is competitive with state-of-the-art baselines.
}

\keywords{Bayesian Optimization, Gaussian Processes, Normalizing Flows}

\maketitle

\section{Introduction}\label{sec1}

Gradient-free global black-box optimization is perhaps the most widely-shared task among scientific and engineering applications.
Such problems are often non-convex, multi-modal, and rugged - presenting a serious challenge for any local optimization strategy \cite{larson_menickelly_wild_2019}.
While there are a host of well-known global methods with which to attack this problem (e.g. population-based methods \cite{Price-et-al-differential-evolution-2005}, gradient-based solvers with finite difference approximations \cite{NoceWrig06}, evolutionary algorithms \cite{DBLP:conf/gecco/Jong19}, simplex-based methods \cite{10.1093/comjnl/7.4.308}), alternatives must be considered in the limit of very expensive objective function evaluations.
By comparison, active learning (AL) \cite{pmlr-v16-settles11a} refers to a class of methods suited to problems where objective evaluation is expensive and requires the next point in the input parameter space to be very carefully chosen.
A popular form of active learning is Bayesian Optimization (BO), which has demonstrated strong performance in the limit of expensive objective evaluation (which translates into a small function call budget for fixed wall-clock time) \cite{ShahriariBobak2016TtHO,2010arXiv1012.2599B,2012arXiv1206.2944S}. 
BO is a probabilistic method that iterates over steps consisting of two components: 1. fitting a surrogate model to the objective function and 2. selecting new points at which to evaluate the expensive objective (via an ``acquisition function'') - the goal is to strike a balance between exploitation and exploration.
The details of BO then boil down to what kind of surrogate model and acquisition function (AF) are chosen.

Gaussian Processes (GPs) are almost always the core of BO algorithms due to their analytic form and interpolation properties \cite{2018arXiv180702811F}.
GPs provide both a surrogate model, through the GP mean $\mu(\theta)$, as well as, crucially, an estimate of uncertainty at unseen points, through the GP error $\sigma(\theta)$ \cite{2006gpml.book.....R}.
The uncertainty estimate is governed by the form of the GP kernel and its hyperparameters, which effectively provide a smoothness prior on the objective function, and is necessary to prioritize exploration to avoid getting stuck in local optima.
GP uncertainties, while theoretically compelling, are not the only tool for exploration in the context of active learning.
Due partially to the poor scaling of linear algebra operations necessary for GP fitting and evaluation, and partially due to the related GP issues in high-dimensional problems \cite{2021arXiv211105040B}, BO-like AL methods have been proposed with alternative surrogates.
Such alternatives include methods that use neural networks (NN) \cite{2015arXiv150205700S,NIPS2016_a96d3afe}, random forests \cite{10.1007/978-3-642-25566-3_40}, or radial basis functions \cite{doi:10.1080/0305215X.2012.687731} as surrogate models.
However, to produce an exploration strategy in the absence of GP uncertainty, such models employ strategies such as model ensembling to estimate uncertainty.
Ensembling is often empirically effective, but can lead to overconfident uncertainty estimates and can be computationally expensive for deep models like NNs \cite{2020arXiv200613570W, 2021arXiv210213640H, 2020arXiv200206470A,2021arXiv210607998M}.

In this contribution 
we argue that there is another way 
to estimate the uncertainty of the 
surrogate model, which is to use  
density estimation: if the local density 
of the sampling points is relatively high in the parameter space of interest, then the
surrogate uncertainty will be lower than other locations in this space, 
and vice versa. 
We propose to use normalizing flow (NF) density estimation of evaluated points to estimate this uncertainty.
This allows us to develop an exploration strategy using the NF density estimate by constructing a novel acquisition function, which is theoretically motivated by deterministic Langevin dynamics.
We demonstrate that this acquisition function, when combined with a local-global hybrid strategy similar to that recently proposed for the trust-region BO method of \cite{2019arXiv191001739E}, meets or exceeds the performance of this method as well as that of widely-used evolutionary and finite-difference gradient algorithms on standard test functions and objectives relevant for probabilistic inference in science and engineering.

\section{Deterministic Langevin Optimization}
\label{sec:method}

We will phrase all discussion in terms of maximization of a scalar objective $f(\theta)$, input parameters $\theta \in \mathbb{R}^d$, and in the context of the problem $\theta^* \in \arg_{\theta} \max f(\theta)$. 
When the goal of optimization
is minimization we change the sign of $f$. 

\subsection{Theoretical Motivation}

The overdamped Langevin 
equation is a stochastic differential 
equation describing particle 
motion in an external potential and 
subject to a random force with zero mean,
\begin{equation}
  \frac{d\theta}{dt}={v}=\beta \nabla_{\theta} f(\theta)+\eta,
    \label{eqn:Lan3}
\end{equation}
where $\langle {\eta(t)}\rangle=0$ and $\langle \eta_i(t) \eta_j(t') \rangle=2\delta_{ij}\delta(t-t')$ where $\langle\rangle$ denotes the expectation taken wrt. the distribution of $\eta$, $v$ is the velocity of the particles with position $\theta$ (with $v,\theta \in \mathbb{R}^{d}$), 
and $\beta$ is the inverse temperature. 
We set the diffusion coefficient
to unity.

The Langevin equation can be viewed as a particle 
implementation of the evolution of 
the (unnormalized) particle probability density $q(\theta_t)$, which is governed by the deterministic
Fokker-Planck equation, a continuity equation for the density,
\begin{align}
    \frac{dq(\theta_t)}{dt}+\nabla_{\theta}\cdot J=0,\\ {J}=q(\theta_t)\nabla_{\theta}[\beta f(\theta_t)+  V(\theta_t)]\equiv q(\theta_t){v}. 
    \label{eqn:fp}
\end{align}
Here $\theta_t$ is the particle position at a some time $t$ and we defined $V(\theta_t)=-\ln q(\theta_t)$ and 
expressed the current $J$ as density times velocity, 
where the two terms in the probability current $J$ correspond
to the two velocity terms in the Langevin equation. 
When we reach a stationary distribution where $\frac{dq(\theta_t)}{dt}=0$, the corresponding density is given by
$q(\theta) \propto \exp(\beta f(\theta))$. 
Thus the solution of the Fokker-Planck 
equation, combined with a suitable 
temperature annealing, where we 
start with $\beta=0$ and 
end with $\beta \rightarrow \infty$, 
will lead us to the solution of the optimization problem. 

In practice, solving the Fokker-Planck 
equation in high dimensions is 
difficult \cite{2022arXiv220514240G}. Instead, 
if we replace the
stochastic velocity in 
the Langevin equation \ref{eqn:Lan3} 
with the deterministic velocity in 
equation \ref{eqn:fp}, we obtain the
\textit{deterministic Langevin equation} \cite{2020Entrp..22..802M,song2020score}, which in discretized form is
\begin{equation}
    \theta_{t+1}=\theta_t+{v}\epsilon=\theta_t+\nabla_{\theta}[\beta f(\theta_t)+  V(\theta_t)]\epsilon,
    \label{Lan4}
\end{equation}
where $\epsilon$ is the step size 
and $V_t$ is the negative
logarithm of the density defined by 
the positions $\{\theta_n\}$ of all previous evaluations. The first term in brackets on the right-hand side of this equation moves
the particles in the direction of 
the target peak $\beta f(\theta)$, 
while the second term moves them 
in the direction of low density as
defined by the points that have been previously evaluated. 
Equation \ref{Lan4} can thus be interpreted heuristically as a 
gradient-based maximization of the objective $\beta f(\theta)+V_t(\theta)$, or
\begin{equation}
   \theta_{t+1} \in \arg \max_\theta [\beta f(\theta)+V(\theta)]  = \arg \max_\theta \ln  \frac{\exp(\beta f(\theta))}{q_t(\theta)}. 
    \label{UV}
\end{equation}
This equation makes manifest the exploration versus 
exploitation nature of the deterministic 
Langevin objective: we 
can either move into the region of the highest
$\exp(\beta f(\theta))$ (exploitation), or 
we can move into the region of the lowest $q_t(\theta)$, which 
we may not have explored yet 
(exploration). 
It is not our intention to mimic the diffusion process of the Fokker-Planck or Deterministic Langevin equations, but to motivate the placement of the next sampling point using the largest discrepancy between the current density and target density.
This target objective is time index $t$-dependent: as we explore a region by evaluating particles in it, its density increases and this reduces the objective at that position. Furthermore, we can also adjust the temperature annealing such that the target objective $\exp(\beta f(\theta))$ is shallow initially, and peaked at the end. 

\subsection{Surrogate and DLO acquisition function}

To make optimization of the objective of equation \ref{UV} into a working algorithm 
we still need several ingredients. 
First, we need to be able to propose values of $\theta$ based on current knowledge of $f(\theta)$ via an acquisition function, which requires a surrogate model.
For the surrogate model $s(\theta)$ of the target $f(\theta)$ in the acquisition function, we deploy Gaussian Processes (GPs) in the main results of this work, 
as the GP is the standard surrogate choice
in Bayesian Optimization algorithms. 
However, we do not take advantage of 
the GP error estimate, so other 
surrogates such as 
neural networks (NN) \cite{2015arXiv150205700S,NIPS2016_a96d3afe}, random forests \cite{10.1007/978-3-642-25566-3_40}, or radial basis functions \cite{doi:10.1080/0305215X.2012.687731} could also
be used. 
We provide a limited discussion extending our results, which are presented for only moderate dimension, to higher dimension using simple neural network surrogates in Appendix~\ref{app:change_surrogate}.
The other ingredient of the acquisition function is the uncertainty estimate, $q(\theta)$, which we return to in Section~\ref{subsec:NFs}.

We incorporate the surrogate into the DLO objective equation \ref{UV} by employing the following acquisition function
\begin{equation}
    {\rm DLO}(\mathbf{\theta};\beta) \equiv  s(\theta;\beta) - \ln q_t(\theta) =\ln
    \frac{e^{s(\mathbf{\theta};\beta)}}{q_t(\mathbf{\theta})} 
    .
    \label{eqn:AF}
\end{equation}
Here the surrogate is fitted to the annealed target as $s(\theta;\beta) = \mathcal{GP}(\{\theta\},\{\beta\times f(\theta)\})$ - i.e., the GP takes as arguments the current ``data'' given by $\{\theta\},\{\beta\times f(\theta)\}$ pairs.
In our default implementation of DLO, $s(\theta)$ is a GP mean, while $q_t$ is an NF fitted to the sample density
after $t$ sample evaluations.

\subsection{Normalizing Flows}
\label{subsec:NFs}

The main difficulty in solving equation \ref{UV} is in evaluating the instantaneous density 
term $q_t(\theta)$. Here we address this by
by feeding all previously evaluated 
points $\theta_1,...\theta_t$
into a normalizing flow (NF) to evaluate the  density of these samples. 

\begin{figure}[h!]
  \centering
  \includegraphics[width=0.85\columnwidth]{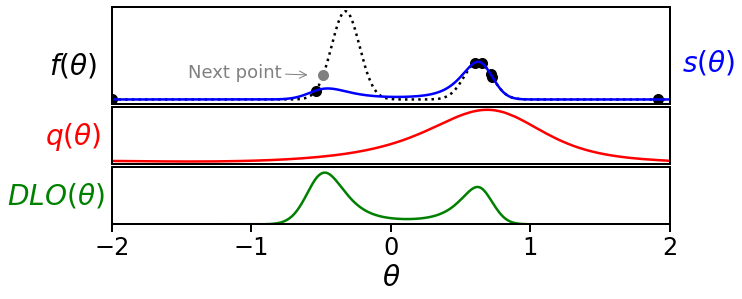}
  \caption{A schematic depiction of the Deterministic Langevin Optimization algorithm.
  The key elements of the algorithm are shown from top to bottom at an early ($\beta_i$) stage of the algorithm (with 7 calls) applied to a mixture of Gaussians.
  \textit{Top:} The surrogate model $s(\theta,\beta_i)$ (solid) begins to fit the target objective (black dotted).
  \textit{Center:} The density estimate $q(\theta)$ gives a smooth  density estimate.
  \textit{Bottom:} The acquisition function $DLO$ (equation~\ref{eqn:AF}) determines where the next point will be selected (gray point).
  At this stage of DLO, the algorithm has determined that the smaller peak is sufficiently explored (density $q$ is high) and now turns to explore the second local peak of surrogate containing the true maximum. The next iteration following the one in the illustration identifies the true maximum at $\theta=-0.325$.
  \label{fig:schematic}
  }
\end{figure}

Normalizing flows are deep generative models that provide a map from a simple ``base'' distribution, such as a uniform or standard normal distribution, to a desired target distribution.
Normalizing flows provide a powerful framework for density estimation and sampling \cite{dinh2016density, papamakarios2017masked,kingma2018glow,dai2021sliced}. These models map the $d$-dimensional data $\theta$ to $d$-dimensional latent variables ${z}$ through a sequence of invertible transformations $\Psi = \Psi_1 \circ \Psi_2 \circ ... \circ \Psi_L$, such that ${z} = \Psi(\theta)$ and ${z}$ is mapped to a base distribution $\pi({z})$, which we choose to be a standard Normal distribution $N(0,\bi{I})$. The probability density of data $\theta$ can be evaluated using the change of variables formula:
\begin{align}
    \label{eq:flow}
    q(\theta)\equiv  e^{-V(\theta)} &= \pi(\Psi(\theta)) \vert\det (\frac{\partial \Psi(\theta)}{\partial \theta})\vert \nonumber \\
    &= \pi(\Psi(\theta)) \prod_{l=1}^L \vert\det (\frac{\partial \Psi_l(\theta)}{\partial \theta})\vert .
\end{align}
The Jacobian determinant of each transform $J_{l}= \vert \det (\partial \Psi_{l}(\theta)/\partial \theta) \vert $ must be easy to compute in order to evaluate the density, and the transformation $\Psi_l$ should be easy to invert for efficient sampling.
In this paper we use the Sliced Iterative Normalizing Flow (SINF) \cite{dai2021sliced} NF algorithm.
SINF scales well to high dimensions and shows good performance on small training sets while using a low number of hyperparameters.
The details of SINF 
are discussed in the Appendix~\ref{app:DLO_hyper}.

\subsection{Local Exploration}
\label{subsec:local_exploration}

To optimize the DLO objective of equation \ref{eqn:AF} we could use gradient-based optimization, since both the GP and NF are differentiable. 
However, this local strategy will be susceptible to getting stuck in local extrema.
Instead, we use a local exploration strategy that is common in BO, drawing a fixed number of proposal samples $N_{\rm sample}$ around the points of higher DLO objective. 
We choose $N_{\rm sample}=100d$.

To generate half of the proposal domain points, we adopt a local exploration strategy similar to \cite{2019arXiv191001739E}, generating domain points within a 
hyperrectangle (a Gaussian sphere also works well)
characterized by a length scale $R$ that grows and shrinks with iteration.
As in \cite{2019arXiv191001739E}, $R$ is initialized to a prescribed starting value and grows and shrinks by a logarithmic step $dR$ depending on how many iterations have passed since $\Delta f_i >0$.

The bijective map of the normalizing flow permits efficient sampling of the target distribution.
We make use of this capability by sampling the other half of proposal samples at each iteration from the latent space of the density estimation normalizing flow $q_{t}(\theta)$.
This is especially helpful for sharply-peaked high-dimensional objectives, as we avoid considering proposals in large regions of essentially zero objective value (even within our local proposal volume).
We draw half of the proposal samples from a Gaussian sphere of radius $R$ in the latent space of the normalizing flow  centered on $\theta^{*}$ after it has been mapped into latent space.
Once the proposal points are 
chosen we evaluate their DLO objective
of equation \ref{eqn:AF} and 
choose the highest value. 
For more details on the effect of removing local exploration from DLO when applied to test objectives, see Appendix~\ref{app:nobeta_nobox}.

\subsection{Simulated Annealing}
\label{subsec:annealing}

We employ simulated annealing \cite{1983Sci...220..671K} to reduce the scale of variation of our target objective early on in the DLO iteration procedure.
The surrogate model fitted to the annealed posterior is denoted $s(\theta,\beta)$, where $\beta$ is the annealing parameter.
It is well-known that simulated annealing is beneficial for optimizing rugged or multi-modal objectives, and is also valuable for population-based Bayesian sampling strategies applied to such challenging functions \cite{Doucet2001}.
We want to design an annealing 
scheme, where we vary $\beta$ between initial and 
final values $\beta_0$ and $\beta_{\rm max}$, such 
that we achieve a good simultaneous exploration 
and exploitation strategy. 
We set the inverse temperature ($\beta$) simulated annealing schedule using $N_{\beta}$ logarithmically-spaced steps in $\beta$, where $N_{\beta}$ is determined by the objective function call budget $N$ and batch size $B$ (with $B$ supplied by the user based on computational constraints), 
$N_\beta=(N-N_I)/B$, where $N_I$ is the number of 
initial samples. 

In our experiments we observe that there is no single value of $\beta_{\rm max}$ that gives the best performance but on many examples $\beta_{\rm max}=100$ and $\beta_{\rm max} \rightarrow \infty$ perform well. 
The latter ignores NF: this is a good strategy when the objective function is isotropic and broad. 
In this case, NF density estimation does not help much with the optimization strategy. 
In other situations, specifically for posterior objectives where large fractions of the input space have very low density, NF exploration is essential for optimal results. 
We note that using only $\beta_{\rm max}=100$ still leads to competitive performance. 
As a result, we will choose between these two values based on their performance, starting with an annealing scheme that combines them with a 50-50 split, which can be adjusted during the annealing if needed according to their performance. 

We also need to set the initial annealing level $\beta_0$. 
We select $\beta_0$ such that the difference 
between the largest and smallest value of $f(\theta)$ on $N_I=2d$ initial samples is less than 15. This ensures sufficiently smooth GP interpolation of initial samples such that early on, if one initial function value dominates the rest by several orders of magnitude, the smaller values are not ignored if they are non-zero (this is especially beneficial for very narrow high-dimensional objectives).
Choosing $\beta_0 \neq 1$ performs a similar function as standardizing the objective function output (as is done, e.g., in \cite{2019arXiv191001739E} using an affine transformation).

Sometimes the $\beta_{0}$ condition is already satisfied for $\beta_{\rm max}=100$. 
In this case we do not anneal at all. 
For all of the synthetic objectives in Section~\ref{subsec:synthetic_objectives}, the range of objective function variation is small enough that annealing is not needed.
However, for posterior objectives and hyper-parameter optimization (HPO) the annealing schedule is essential for good performance.
In practice, there is a broad range of $\beta_0$ values that give the same performance. 
For more details on the effect of removing the simulated annealing step from DLO when applied to test objectives, see Appendix~\ref{app:nobeta_nobox}.

\subsection{Algorithm}
\label{subsec:algorithm}

The high-level Deterministic Langevin Optimization algorithm is presented in Algorithm~\ref{alg}. 
We rescale the domain of $\theta$ to the product of $d$ unit intervals.
Each update $\theta_{t}\to\theta_{t+1}$ contains a batch of $B$ samples.

\begin{algorithm}[H]
\caption{Deterministic Langevin Optimization}\label{alg}
\begin{algorithmic}[1]
\State Evaluate $f(\theta_{1}),..,f(\theta_{N_I})$ at $N_I$ initial points; select initial annealing level $\beta_0$, rescale the input $\theta$ domain to $[0,1]^{d}$. 
\State Assign a call budget $N$, fix the hyperparameters $N_{\beta}$, $R$, $dR$.
\For{$i < N_{\beta}$}
 \State Estimate the normalizing flow density $q_{i}(\theta)$ from $\theta_1,..,\theta_t$.
 \State Fit the surrogate $s_i(\theta,\beta_i))$ from $f(\theta_{1}),..f(\theta_{t})$ to annealed objective values.
 \State Create proposal samples in $[0,1]^d$ \textit{and} in the latent space of $q_{t}$ drawing from Gaussian spheres of radius $R$ around the highest ${\rm DLO}(\theta_j),j=1...t$.
 \State Locally maximize the acquisition function ${\rm DLO}(\theta)$ from $N_{\rm sample}$ proposal draws to obtain the next batch of $\theta_{t+1},..,\theta_{t+B}$ to evaluate.
 \State Evaluate $f(\theta_{t+1}),..f(\theta_{t+B})$ and update $\beta_i$.
 \EndFor
 \end{algorithmic}
 \end{algorithm}

\section{Numerical Experiments}
\label{sec:numerical_experiments}

We test DLO on three types of test objectives in this section.
We first show performance on several standard gradient-free optimization test functions, before turning to posterior objectives relevant for applying DLO to Bayesian posterior optimization relevant for inference in science and engineering, and finally demonstrate that DLO is competitive on several applied objectives, including Hyperparameter Optimization (HPO). 
We also 
compare our new aquisition function eqn.~\ref{eqn:AF} to several alternatives that hew closer to the literature.

We compare DLO to \texttt{TuRBO}, which, for general-purpose single-objective gradient-free optimization, is an extremely popular highly-performing BO method.
The local optimization aspect of our method is based on a trust region strategy \cite{2019arXiv191001739E} - so this comparison is especially relevant for illustrating the improvement furnished by our modified acquisition function.
We use the \texttt{TuRBO} hyperparameters and numerical setups as detailed in \cite{2019arXiv191001739E} unless otherwise stated.
We further compare to differential evolution (DiffEv) \cite{Price-et-al-differential-evolution-2005} and the covariance matrix adaptation evolution strategy (CMA-ES) \cite{2016arXiv160400772H}, which give an indication of how population-based and evolutionary strategies perform in the limit of a low number of function evaluations, and to the Broyden-Fletcher-Goldfarb-Shanno (BFGS) algorithm with finite difference gradients.
We also include a random search baseline.
For all experiments we consider only sequential evaluation (batch size of 1) function evaluations for DLO unless otherwise noted.
Details of the numerical experiments are provided in Appendix~\ref{app:details}.

\subsection{Synthetic Objectives}
\label{subsec:synthetic_objectives}

We consider two common synthetic test functions in the gradient-free optimization literature in $10$ dimensions - the Ackley function and Rastrigin function.
Both functions are characterized by the presence of many local optima, with the Ackley function having a global optimum that is narrow and much higher than the local minima, while the Rastrigin function has a shallow global curvature with more pronounced local optima.
We use domains of $[-5,10]^{10}$ and $[-5.12,5.12]^{d}$ for the Ackley and Rastrigin functions, respectively.
These functions are usually given as minimization objectives.
Since our method is formulated with posterior inference applications in mind, we transform these synthetic objectives into maximization problems as $g_{\rm{min}} \to g_{\rm{max}} \equiv \log\left(g(\theta^*)-g(\theta)\right)$.

\begin{figure}[h!]
  \centering
  \includegraphics[width=0.46\columnwidth]{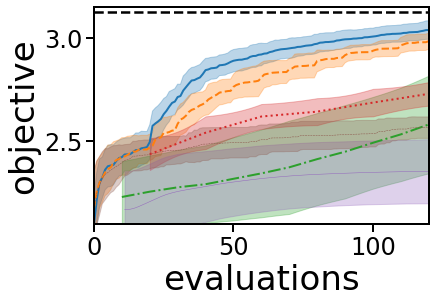}
  \includegraphics[width=0.45\columnwidth]{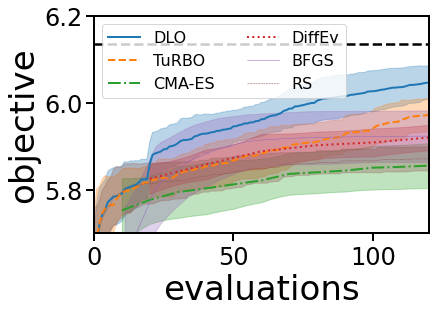}
  \caption{Performance of DLO relative to other methods on converging to the optimum on $10-d$ synthetic objectives. \textit{Left:} Ackley function. \textit{Right:} Rastrigin function.}
  \label{fig:synthetic}
\end{figure}

Results on these functions are presented in Figure.~\ref{fig:synthetic}, where we see that DLO outperforms the other methods in the low function call budget regime (over 30 optimization runs).
On both objectives, DLO sees an extra jump in performance early on in the optimization runs with respect to the other methods, which can be attributed to our modified acquisition function.
The evolutionary methods and BFGS, meanwhile, are not competitive with DLO and \texttt{TuRBO} on these functions.

\subsection{Posterior Objectives}
\label{subsec:posterior_objectives}

\begin{figure}[h!]
  \centering
  \includegraphics[width=0.46\columnwidth]{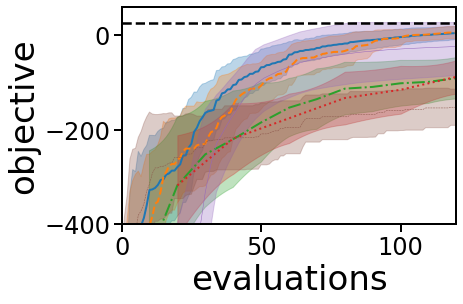}
  \includegraphics[width=0.45\columnwidth]{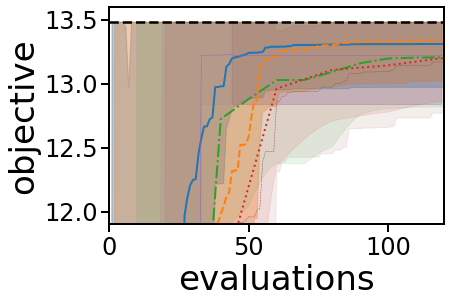}
  \includegraphics[width=0.45\columnwidth]{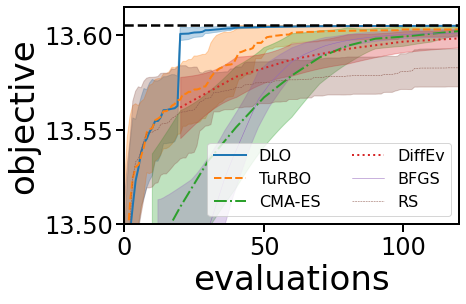}
  \caption{Performance of DLO relative to other methods on converging to the optimum on $10d$ posterior objectives. \textit{Left}: Correlated Gaussian with condition number 200. \textit{Right}: Widely-separated double Gaussian mixture. \textit{Bottom}: Rosenbrock posterior.}
  \label{fig:posterior}
\end{figure}

A key application of AL methods is to expensive scientific inference problems \cite{2019arXiv191101429C}.
Such problems (e.g. physical simulations \cite{2019JCAP...02..031R}) may require model evaluations that take minutes, hours, or days of wall-clock time on multiple nodes of high-performance computing facilities. 
Efficient parameter estimation with as few samples as possible is thus of paramount importance, and MAP optimization is faster than Monte Carlo Markov Chains, which are often 
prohibitively expensive.
With this wide class of applications in mind, we consider several 
10-dimensional test posterior objectives of the type that frequently arise in scientific parametric inference applications.

A correlated Gaussian distribution is the simplest non-trivial posterior routinely encountered in statistical inference.
We make this problem challenging by enforcing a high condition number (ratio of maximum to minimum eigenvalues) of 200 in the test posterior.
Figure~\ref{fig:posterior} shows that DLO (and \texttt{TuRBO}) acheive the best performance, and are consistent with each other, though DLO has a slight edge early-on.
BFGS also attains decent performance, but with a large variance between runs.

Posteriors that resemble Gaussian mixtures arise frequently in scientific inference (e.g. in inferring gravitational wave merger parameters \cite{2016PhRvL.116x1102A}).
We consider a widely-separated double Gaussian posterior with thin peaks that has a false maximum (one peak is higher than the other, and contains 70\% of the probability mass).
While the difficulty of this example produces a large variance between runs (especially for the evolutionary algorithms), Figure~\ref{fig:posterior} shows that DLO typically finds a point near the true maximum first, followed by \texttt{TuRBO}, and eventually the evolutionary algorithms, while later-evaluation performance of DLO and \texttt{TuRBO} is similar.
For this example we note that the gradient-based BFGS typically converges first  - but this is to the wrong peak 9 times out of 30, and unlike the other algorithms, has no chance at improving after this point.

We also consider the non-convex Rosenbrock function as a posterior (after suitable shifting to ensure a positive objective), which is a test frequently considered by Bayesian sampling methods \cite{2015MNRAS.450L..61H} due to thin curving posteriors that frequently arise in scientific applications.
We find that DLO attains the best performance on this objective by far, jumping up to the optimum before \texttt{TuRBO} and all other methods (including BFGS).

\subsection{Applied Objectives}
\label{subsec:applied_objectives}

\begin{figure}[h!]
  \centering
  \includegraphics[width=0.5\columnwidth]{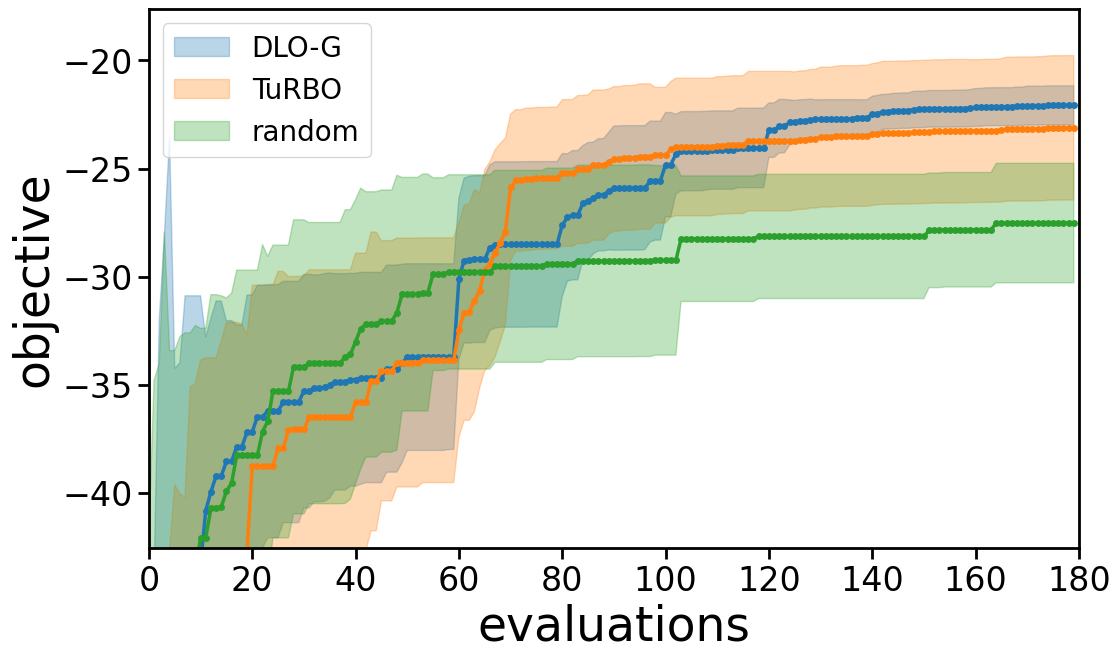}
  \includegraphics[width=0.4\columnwidth]{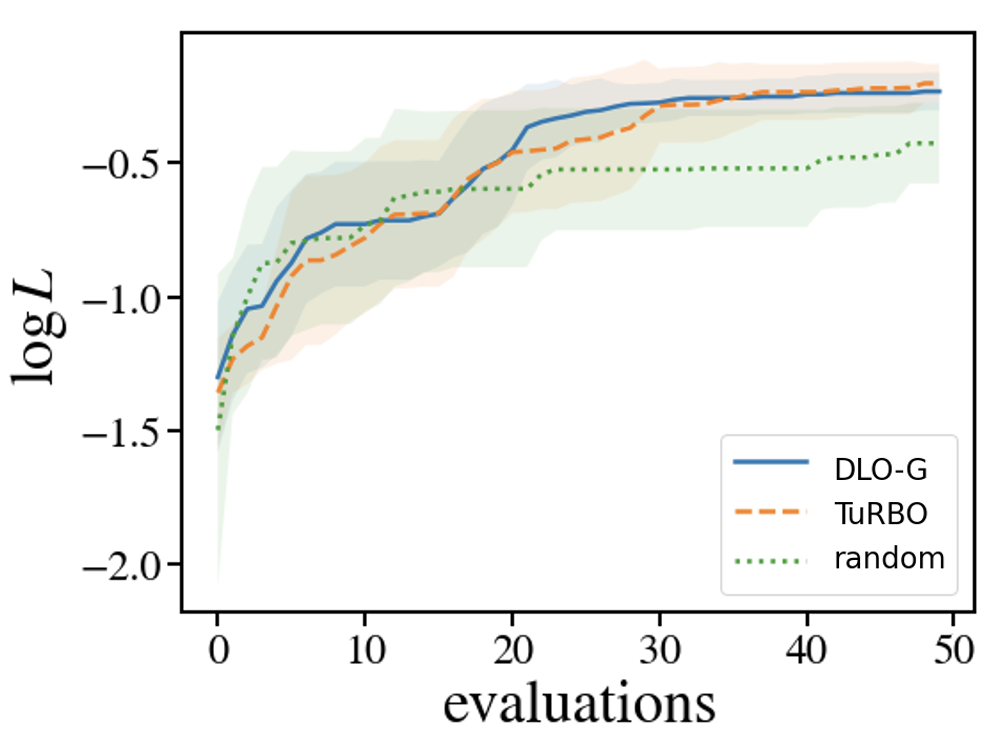}
  \caption{Performance of DLO compared to two baselines (\texttt{TuRBO} and random search) on converging to the optimum on several applied objectives. \textit{Left}: ``Cosmological Constants'' posterior  \textit{Right}: Neural network (Multi-layer Perceptron) HPO (blue line corresponds to DLO). }
  \label{fig:applied}
\end{figure}

\subsubsection{``Cosmological Constants''}

The ``Cosmological Constants'' problem is the optimization of a $12d$ posterior objective relevant for the analysis of cosmological parameters from Sloan Digital Sky Survey Luminous Red Galaxy clustering data \cite{2010MNRAS.404...60R}. 
The dominant cost of the posterior evaluation is the solution of a large system of stiff differential equations - the Einstein-Boltzmann equations.
For this problem we fix the neutrino mass to zero, which greatly reduces the time-to-solution of the Einstein-Boltzmann solver, but leave all other parameters free.
Figure~\ref{fig:applied} illustrates that DLO outperforms TuRBO (averaged over 15 runs).

\subsubsection{Machine Learning Model Hyperparameter Optimization}

BO-style AL methods are well-suited to hyperparameter optimization (HPO) for machine learning models, and have begun to gain traction over random/grid search among practitioners \cite{wandb,2019arXiv190710902A,2021arXiv210906716E,2021arXiv210909831L} since they outperform such simple methods \cite{2021arXiv210410201T,2019arXiv191011858W,2018arXiv180609055L,2021arXiv210402487S}.
In particular, the \texttt{TuRBO} baseline and its competitor-supplied variants performed well in the 2020 Bayesian Black-Box optimization challenge for several simple models on publicly available (scikit learn) datasets.
We consider an example drawn from this challenge, in particular we use the \texttt{bayesmark}\footnote{\url{https://bayesmark.readthedocs.io/}} experiment for multi-layer perceptron SGD applied to the  \texttt{scikit-learn} \cite{scikit-learn} ``iris'' dataset with the (negative) log likelihood objective, in Figure~\ref{fig:applied}. This
Figure shows that DLO is competitive with \texttt{TuRBO} (over 15 replications) on this HPO problem, and gives results that are consistent with \texttt{TuRBO} in this small-budget setting.

\subsection{Modifying the acquisition function}
\label{subsec:change_AF}

\begin{figure}[h!]
  \centering
  \includegraphics[width=0.75\columnwidth]{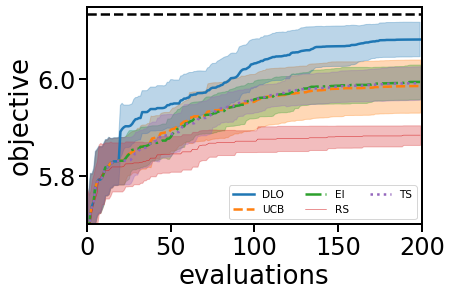}
  \caption{
  Performance of the DLO algorithm for several choices of standard GP acquisition function on the $10d$ Rastrigin function. DLO shows a clear improvement over the other acquisition strategies.
  }
  \label{fig:ablation}
\end{figure}

We perform an acquisition function ablation study to characterize the extent to which the NF density estimate equation~\ref{eqn:AF} is responsible for the strong performance of DLO shown in Sections~\ref{subsec:synthetic_objectives}-\ref{subsec:applied_objectives}.
Figure~\ref{fig:ablation} shows progress toward the optimum of DLO when applied to the $10d$ Rastrigin function for several common BO acquisition functions: Expected Improvement (EI) \cite{jones}, Upper Confidence Bound (UCB) \cite{6138914}, and Thompson Sampling (TS) \cite{10.1093/biomet/25.3-4.285,2017arXiv170702038R}.
These AFs all incorporate the GP uncertainty estimate to guide the extent to which to prioritize exploration, and, for reference, are given by:
\begin{align}
    {\rm EI}(\theta) &= (\mu(\theta) - f^*(\theta))\Phi\left(\frac{(\mu(\theta) - f^*(\theta))}{\sigma(\theta)}\right)\nonumber \\ &+ \sigma(\theta)\phi\left(\frac{(\mu(\theta) - f^*(\theta))}{\sigma(\theta)}\right) \\
    {\rm UCB}(\theta) &= \mu(\theta) + \beta_{\rm UCB}\sigma(\theta),\\
    {\rm TS}(\theta) &= f\sim \mathcal{GP}(\mu(\theta),\sigma(\theta)).
\end{align}

and we set $\beta_{\rm UCB}=1$.
Figure~\ref{fig:ablation} shows that DLO (with $\beta_{\rm{max}}=100$) has a distinct edge over the other acquisition functions, especially very early on in the optimization runs.

\section{Related Work}
\label{sec:related_work}

Gradient-free optimization is one of the most active areas of 
global optimization
and there is a correspondingly vast literature covering many methods. We review some aspects of these methods that are most relevant to DLO.

\textbf{Local+global methods:}
The \texttt{TuRBO} strategy \cite{2019arXiv191001739E} has led to many follow-up methods (e.g. \cite{2021arXiv210703217M}) and applications  - including in constrained problems \cite{2020arXiv200208526E}, BO with axis-aligned features \cite{2021arXiv210300349E}, and multi-objective problems \cite{2021arXiv210910964D}.
Basin hopping has long been used by the chemistry community for optimizing multi-modal functions \cite{1998cond.mat..3344W}, and other global BO methods with local BFGS searches have recently been developed for atomic structures \cite{2019PhRvB.100j4103G}. 
The DLO global strategy differs from these, while it can still take advantage of the many possible local methods proposed in the literature. 

\textbf{Alternative uncertainty estimates:} 
Several studies have explored supplying a different surrogate model in place of a GP, which must come with a different uncertainty estimate.
Alternatives that have been explored include random forests (\texttt{SMAC} \cite{10.1007/978-3-642-25566-3_40}), neural networks (\texttt{DNGO} \cite{2015arXiv150205700S}, \texttt{BOHAMIANN} \cite{NIPS2016_a96d3afe}).
While such surrogates differ from the standard GP, these methods have retained versions of (\cite{mockus2012bayesian}) Expected Improvement (EI) as their acquisition functions by estimating the variance a GP would have provided from these surrogate models (i.e. by looking at variation over RF trees or via a Bayesian last-layer treatment). 
\texttt{NOMU} \cite{2021arXiv210213640H} incorporates a more complex NN uncertainty estimate by training two dependent NNs and using a carefully constructed (non-EI) loss function.
DLO replaces
uncertainty quantification with 
density estimation: the density of samples in some region is lower, it is expected that the interpolation quality of the surrogate will be lower, which we can interpret as higher surrogate uncertainty. 

\textbf{Modified acquisition functions:}
Ref.~\cite{2017arXiv170400520J} proposed an acquisition function that places increased weight on BO exploration with the goal of Bayesian posterior approximation.
This was employed in a BO framework applied to scientific simulators \cite{JMLR:v17:15-017,2018PhRvD..98f3511L}, but was restricted to relatively low-dimensional examples.
Other recent work \cite{2015arXiv151006299G,pmlr-v108-yue20b,pmlr-v108-yue20b,Wu2019PracticalTL} has explored non-myopic acquisition functions, which make the choice of which point to evaluate based on predictions of future evaluations.
These predictions are made by (approximately) integrating over future GP predictions, and therefore employ GP uncertainty.
The \texttt{BORE} method \cite{2021arXiv210209009T,2022arXiv220316912D} employs density-ratio estimation as part of its acquisition strategy in the context of formulating EI as a classification task, which is distinct from our evaluated sample density estimation. DLO AF differs
from all of these methods, since it uses 
NF density estimation as part of AF.

\section{Conclusions}
\label{sec:conclusion}

Probabilistic surrogate models are a powerful component of gradient-free black-box global optimization.
While variants of Gaussian Process-based Bayesian Optimization have demonstrated enormous success on this class of tasks, using a Gaussian Process-based acquisition strategy is not the only competitive choice.
We have demonstrated that Normalizing Flow density estimation provides an exploration strategy that is competitive, and frequently superior, to that provided by GP-based methods in the context of BO-like active learning optimization when incorporated into the Deterministic Langevin acquisition function we propose in this paper.
We demonstrated this through the optimization of several standard optimization test functions, common posterior objectives arising in Bayesian inference, and real-world application objectives, such as cosmological inference and ML hyperparameter optimization. 

The proposed method is not without limitations. One is the choice of the hyperparameter $\beta_{\rm max}$ and the annealing schedule. We do not expect that one choice will fit all problems, a consequence of the no free-lunch theorem of global optimization: in some problems we can use a greedy strategy and progress quickly towards the peak without much penalty, in others we need more exploration to find all the existing peaks. This can be controlled by the choice of $\beta_{\rm max}$.  
In this paper we compare the performance between two different values of $\beta_{\rm max}$, but there may be better ways to do this. 

Another limitation is that the wall-clock time of DLO is currently limited by the $\mathcal{O}(n^3)$ scaling of evaluation of the default Exact GP we employ in \texttt{gpytorch} \cite{gardner2018gpytorch}. 
This may be alleviated through GPU-based methods for accelerating GP kernel operations when many samples are present \cite{2019arXiv190308114W} (for the experiments presented here were run on a single 2.8 GHz Quad-Core Intel Core i7 CPU), or through the use of a more efficient surrogate (we briefly explore this in Appendix~\ref{app:change_surrogate}).

\backmatter

\bmhead{Acknowledgments}
We thank Biwei Dai for assisting with SINF.
We thank Richard Grumitt, David Nabergoj, and Juliane Mueller for useful discussions, and also thank Juliane Mueller for comments on a draft version of this work.
We thank Phil Marcus and Haris Moazam Sheikh for helpful conversations.
JMS was supported by a U.S. Department of Energy Computational Graduate Sciences Fellowship under Award Number DE-SC0019323, and a Department of Energy Office of Science Graduate Student Research Award while this work was conducted.

\section*{Declarations}

\textbf{Data Availability:} 
A public repository of our implementation of DLO and the experiments run here will be made available upon publication via a link provided in this article.

\begin{appendices}

\section{Setup of Numerical Experiments}
\label{app:details}

For all experimental results, we use $N_0 = 2d$ initialization points for each numerical experiment for DLO and \texttt{TuRBO} sampled from a Latin hypercube (except when the implementation of the algorithms, such as for BFGS ad CMA-ES, only permit a single initial value, which is generated from a uniform distribution on the input space in that case).
We use a total function call budget of $N_0+10d$ for each problem, except in Section~\ref{subsec:change_AF}, where we use twice this budget.
We use the \texttt{scipy} implementation of differential evolution and the CMA-ES implementation of \cite{nikolaus_hansen_2022_6370326}.
Though this comparison was already performed in \cite{2019arXiv191001739E}, we also include the \texttt{scipy} \cite{2020SciPy-NMeth} implementation of the Broyden-Fletcher-Goldfarb-Shanno (\texttt{BFGS}) algorithm \cite{NoceWrig06} to demonstrate the challenges gradient-based methods face in the low-budget regime on the objectives we consider.
If BFGS converges before the function call budget is reached, we set the rest of the values in the experiment to the converged value.

The Ackley and Rastrigin functions are given (in 10 dimensions) by
\begin{align*}
    F(\theta) &= 20\left(1 - \exp\left[-\frac15 \sqrt{\frac{1}{10} \sum_i^{10} \theta_i^2}\right] \right) \\&+ e\left(1 - \exp\left[\frac{1}{10}  \sum_i^{10} \cos\left(2\pi\theta_i\right) \right]\right),\\
    F(\theta) &= 100 + \sum_i^{10}\left(\theta_i^2 -10\cos(2\pi\theta_i)\right),
\end{align*}
respectively.
Their domains are $[-5,10]^{10}$ and $[-5.12,5.12]^{10}$.

The posterior objectives we consider are the correlated Gaussian posterior, which is simply a Gaussian distribution with mean $\mu = \frac15\mathbf{u}$, where $\mathbf{u}$ is the $10d$ vector whose entries are all ones, and covariance $C$ such that for the singular values $\lambda_{i}$ of C, $\frac{\lambda_{1}}{\lambda_{10}} = 200$, with $\lambda_{1} = 0.09$.

The double Gaussian posterior is the mixture
\begin{align*}
    F(\theta) &= 0.3 p_N(\theta;0.625,0.1) \\&+ 0.7 p_N(\theta;-0.325,0.1)
\end{align*}
where $p_N(\theta;\mu,\sigma)$ is the multi-variate Gaussian probability density.
We restrict to the domain $[-2,2]^{10}$ for both the correlated and double Gaussian posteriors.
The Rosenbrock posterior is given by
\begin{align*}
    F(\theta) = \sum_{i=0}^{9} (1-\theta_{i})^2 + 100\left(\theta_{i+1} - \theta_{i}^2\right)^2
\end{align*}
on the domain $[-5,5]^{10}$.

The ``Cosmological constants'' example is described in the main text - but involves the maximization of an $11d$ log likelihood for the galaxy power spectrum - a compressed statistic of galaxy survey data - to obtain a point estimate of cosmological parameters.
These parameters include the matter density, expansion rate, and various numerical factors that govern the  differential equation solver that serves as part of the model called by the likelihood.

The HPO example we consider uses the ``iris'' dataset and the \texttt{bayesmark} (\texttt{sklearn}) implementation of a Multi-layer Perceptron which has parameters that are optimized through SGD, and uses the (inverted) negative log likelihood as the objective.
The hyperparameters for this model are the number of hidden layers, the learning rate and its initialization, the batch size, the tolerance, the momentum, a regularization parameter (\texttt{alpha}), and learning rate scaling.

\section{DLO Hyperparameters}
\label{app:DLO_hyper}

In this section we discuss the DLO hyperparameters.

The GP surrogate model is implemented in \texttt{pytorch} \cite{NEURIPS2019_9015} and uses the \texttt{gpytorch} \cite{gardner2018gpytorch} ExactGP implementation. 
Following \cite{2019arXiv191001739E}, we use a GP with Mat\'ern-5/2 covariance, but force the GP noise to be smaller (between $10^{-6}-10^{-4}$), and do not standardize the objective function values.
We optimize over the GP hyperparameters using 50 steps of \texttt{ADAM} \cite{2014arXiv1412.6980K}.

\begin{figure}[h!]
  \centering
  \includegraphics[width=0.45\textwidth]{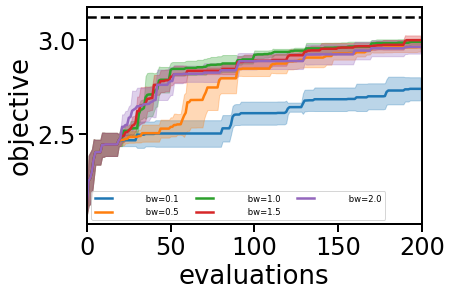}
  \includegraphics[width=0.49\textwidth]{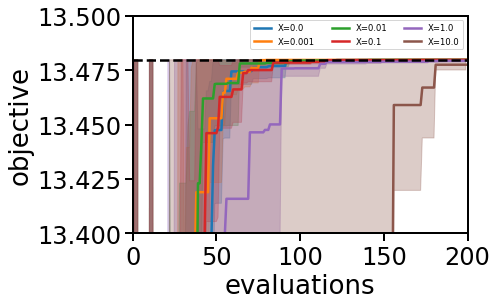}
  \caption{
  \textit{Left:} Dependence of progress toward the optimum on the NF bandwidth factor for the GP-based version of DLO with mean and standard error estimated over 5 realizations.
  \textit{Right:} Dependence of progress toward the optimum on the AF coefficient $X$ for the GP-based version of DLO with mean and standard error estimated over 5 realizations.
  }
  \label{fig:X_bw}
\end{figure}

For the normalizing flow we employ for density estimation (SINF), we use 5 iterations, no regularization and up to 8 sliced Wasserstein directions (the default prescription of \cite{dai2021sliced}). The most important hyperparameter of SINF is a pre-factor $bw$ that modifies a Scott's rule prescription for the SINF density estimation pre-factor (see \cite{dai2021sliced} equation~26).
We demonstrate that variation of this hyperparameter has very little effect on our results in Figure~\ref{fig:X_bw}.
In this study $X=1$, though the $bw$ dependence does not greatly depend on this fact.
Clearly, very small $bw$ (less than 1) leads to somewhat reduced performance, while if $bw$ is ``large enough'' there is no change in performance.
Intuitively this makes sense - a too small $bw$ means that the density estimating flow $q$ will be close to zero except very close to the evaluated objective points, where it will be very high.
This will mean proposals in the support of the density (which is tightly hugging the evaluated points) will have very low AF value.
While a lower AF value in these regions is somewhat desirable (otherwise there will never be any explicit exploration - exploration would only be possible through the implicit error on $s(\theta;\beta)$), if the already-evaluated points are high in the objective, this will slow down convergence to the optimum.
A larger $bw$ effectively regularizes the value of $q$ to avoid the above scenario by softening the penalty near already-evaluated points.
We fix $bw=1.0$ to avoid reducing performance - this is also the default value in SINF.

There is no a-priori way to know how much exploration should be weighted against exploitation.
While the standard acquistion functions $EI$ and $TS$ have no explicit hyperparameters in their functional form (that is, outside of those used in constructing the GP), $UCB$ (which we note performs best in our tests, c.f. Fig.~\ref{fig:ablation}) contains the hyperparameter $\beta_{UCB}$.
While theoretical estimates have been provided for $\beta_{UCB}$ (e.g. in \cite{6138914}), in practice, the same authors choose a value that works better empirically. In our setting 
this is controlled by the value 
$\beta_{\rm max}$. 
We explore the effect of changing $\beta_{\rm max}$ on the double Gaussian example in Figure~\ref{fig:X_bw}. 
Clearly too-large values of $X$ lead to over exploration and slow convergence, while small values show improved convergence.
We select $X=0.01$ as our fiducial value to avoid over-exploring.
The DLO proposal volume evolution is similar to that of \texttt{TuRBO} (which is drawn from the trust region literature).
The volume starts at $R=1$ and decreases by $dR$ if we do not improve (within a tolerance of $5\times10^{-6}$) on the best value $f^*$ after two iterations.
After a successful improvement on $f^*$, the volume increases by the same factor to consider a larger proposal volume.
Rather than using Automatic Relevance Determination to fit an independent GP kernel lengthscale parameter in each coordinate dimension, when not proposing in latent space, we scale the proposal volume along the coordinate axes by the gradient direction unit vector.

\section{Batch size}

\begin{figure}[H]
  \centering
  \includegraphics[width=0.45\textwidth]{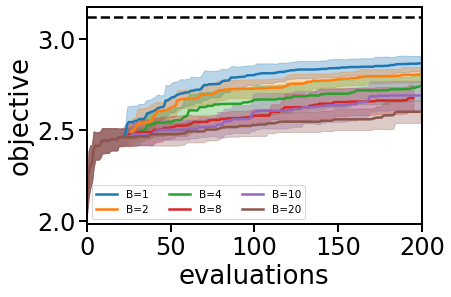}
  \includegraphics[width=0.45\textwidth]{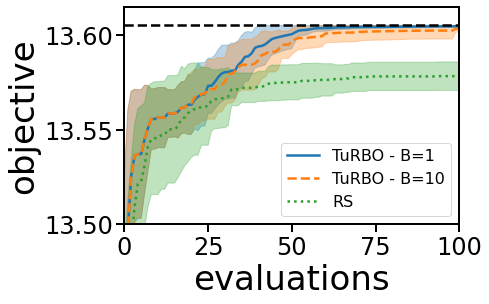}
  \caption{
  \textit{Left:} Dependence of progress toward the optimum for the GP-based version of DLO with mean and standard error estimated over 10 realizations. 
  \textit{Right:} Optimization runs on the $10d$ Rosenbrock function, as in Figure~\ref{fig:posterior}, but changing the \texttt{TuRBO} batch size from 10 (as provided by those authors for $10d$ test functions) to 1.
  We also show a random search baseline.
  }
  \label{fig:batch}
\end{figure}

We show the effect of batch size on our method on the synthetic Ackley test function in Figure~\ref{fig:batch} (here using $X=1$ for all proposals).
A too-large batch leads to somewhat reduced performance, as would be expected due to each batch having less information to use when selecting the next point with the acquisition function.

Figure~\ref{fig:batch} indicates the change in performance of \texttt{TuRBO} when the batch size is reduced to 1 from the values prescribed in \cite{2019arXiv191001739E}.
Following the results presented in Figure~7 of \cite{2019arXiv191001739E}, we see that the performance improves with reduced batch size, but only slightly.

\section{Replacing the surrogate model}
\label{app:change_surrogate}

In the main text, we presented an implementation of the DLO algorithm that employed a GP for the surrogate model and an NF, SINF, for the density estimate.
However, as the number of dimensions of the problem at hand increases, more data points are required to obtain an accurate surrogate model.
Since the cost of the GP surrogate scales as the number of data points cubed, alternative surrogates may be preferred in high dimensions.

\begin{figure}[H]
  \centering
  \includegraphics[width=0.6\columnwidth]{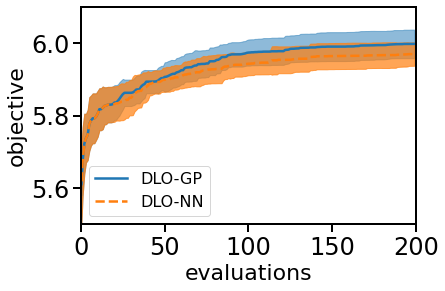}
  \caption{
  Progress toward objective optimum on the 10-$d$ Rastrigin test problem for both a simple neural network surrogate and the default choice of a Gaussian Process surrogate.
  Here we average over 15 realizations and provide the related confidence band.
  }
  \label{fig:surrogate}
\end{figure}

In this spirit, we consider a simple fully-connected neural network surrogate as a replacement for the GP mean surrogate.
In Figure~\ref{fig:surrogate} we use a FCN implemented in \texttt{pytorch} with 2 hidden layers of width 100 trained with ADAM \cite{2014arXiv1412.6980K} and Tanh activation as the surrogate and compare it to the progress toward the optimum on the Rastrigin function in $10d$.
We find slightly reduced performance but a generally comparable result, indicating deep network surrogates can perform well as replacement for GPs.
We have not expended significant time or effort on finding a well-performing architecture, so results can likely be specifically improved upon those shown here, but this is out of the scope of this work, as is further exploration of higher dimensional problems.
However, already it is empirically clear that the NN surrogate is much more scalable than the GP.
Table~\ref{tab:ack_d_time} makes this cost scaling explicit, as while the neural network is always cheaper to fit and evaluate as a surrogate than a GP in the context of DLO, the scaling of the GP cost is significantly worse with dimension for the Ackley objective (here both methods obtain similar performance in terms of the final objective value).

\begin{table}[h!]
    \centering
    \begin{tabular}{|c|c|c|c|c|c|}
    \hline
        $d$ & 2 & 5 & 10 & 20 & 50 \\ \hline
        \multicolumn{6}{|l|}{Evaluation:}\\ \hline
        DLO-GP & 0.02 & 0.03 & 0.07 & 0.15 & 6.07 \\
        DLO-NN & 0.02 & 0.02 & 0.04 & 0.10 & 0.53  \\ \hline 
        \multicolumn{6}{|l|}{Fitting:}\\ \hline
        DLO-GP & 0.17 & 0.26 & 0.44 & 1.71 & 46.65 \\
        DLO-NN & 0.03 & 0.04 & 0.06 & 0.20 & 2.84 \\
    \hline
    \end{tabular}
    \vspace{0.5cm}
    \caption{Surrogate \textit{evaluation} (top rows) and \textit{fitting} (bottom rows) timings (in seconds) for both Gaussian Process (GP) and fully-connected neural network (NN) surrogates at the last iteration of optimization of the Ackely function for several choices of dimension $d$. }
    \label{tab:ack_d_time}
\end{table}

\section{Annealing and local search ablation}
\label{app:nobeta_nobox}

\begin{figure}[H]
  \centering
  \includegraphics[width=0.45\columnwidth]{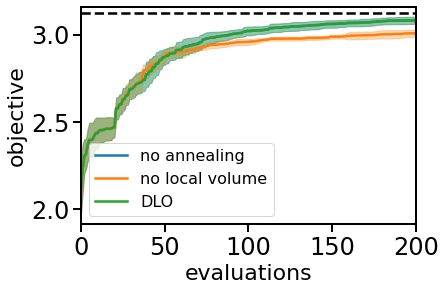}
  \includegraphics[width=0.45\columnwidth]{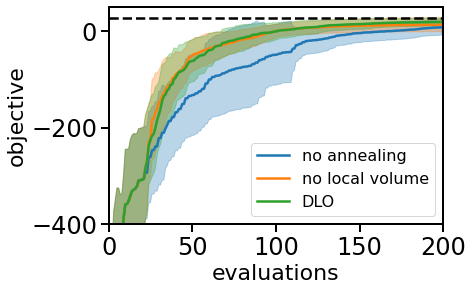}
  \caption{The effect of removing temperature annealing and the local volume search from the DLO algorithm on performance in 10 dimensions for the Ackley function (\textit{left}) and the correlated Gaussian (\textit{right}) objectives (averaged over 15 realizations).
  \label{fig:nobeta_nobox}
  }
\end{figure}

To demonstrate that both the annealing and suggesting candidate points from a local search volume (rather than from a global search) are necessary for the performance shown in the main text, here we perform a simple ablation study and remove these two components of the DLO algorithm.
Figure~\ref{fig:nobeta_nobox} illustrates that for the 10-dimensional objectives, performance suffers when annealing and the local bounding box for choosing candidate points are omitted.
In particular, for the Ackley function, taking out the local volume search gives significantly worse average performance (in terms of objective value obtained), while removing annealing has comparatively little effect.
For the correlated Gaussian posterior, the situation is reversed - removing the local volume search seems to have little effect, while the omission of annealing causes more variation between runs and worse average empirical convergence toward the optimum.
The differences here are not surprising, as the two targets furnish very different optimization problems.
For one, the Gaussian target is convex, and has an enormous dynamic range in objective ($\log p(\theta)$) value, which is dramatically reduced by the temperature annealing. 
Meanwhile, the Ackley function has comparatively little dynamic range, and has many local minima, so it is possible for the acquisition strategy to overfavor far-away local minima if they have not yet been explored when there is no local candidate search volume.

\end{appendices}


\bibliography{sn-bibliography,paper}


\end{document}